\documentclass[lettersize,journal]{IEEEtran}
\usepackage{amsmath,amsfonts,amssymb}
\usepackage{algorithm}
\usepackage{array}
\usepackage[caption=false,font=normalsize,labelfont=sf,textfont=sf]{subfig}
\usepackage{textcomp}
\usepackage{stfloats}
\usepackage{url}
\usepackage{verbatim}
\usepackage{graphicx}
\usepackage{cite}

\usepackage{booktabs} 
\usepackage{boldline}
\usepackage{multirow} 
\usepackage[table]{xcolor} 
\usepackage{colortbl}
\usepackage{xcolor}
\usepackage[colorlinks, linkcolor=blue, citecolor=green]{hyperref}

\usepackage{algpseudocode}

\begin{document}
\bibliographystyle{unsrt}

\title{Cross-modal Offset-guided Dynamic Alignment and Fusion for Weakly Aligned UAV Object Detection}

\author{Zongzhen Liu, Hui Luo, Zhixing Wang, Yuxing Wei, Haorui Zuo, Jianlin Zhang
\thanks{Zongzhen Liu, Hui Luo, Zhixing Wang, Yuxing Wei, Haorui Zuo, Jianlin Zhang are with the State Key Laboratory of Optical Field Manipulation Science and Technology, Chinese Academy of Sciences, also with the Institute of Optics and Electronics, Chinese Academy of Sciences, Chengdu 610209, China. (e-mail: 202244060131@std.uestc.edu.cn; luohui19@mails.ucas.ac.cn; 202111012048@std.uestc.edu.cn; zuohaorui@ioe.ac.cn; yuxing\_wei\_ioe5@163.com; jlin@ioe.ac.cn )} 

\thanks{\textit{Co-corresponding authors: Hui Luo; Jianlin Zhang.}}

}

\markboth{Journal of \LaTeX\ Class Files,~Vol.~14, No.~8, August~2021}%
{Shell \MakeLowercase{\textit{et al.}}: A Sample Article Using IEEEtran.cls for IEEE Journals}


\maketitle

\begin{abstract}

Unmanned aerial vehicle (UAV) object detection plays a vital role in applications such as environmental monitoring and urban security. To improve robustness, recent studies have explored multimodal detection by fusing visible (RGB) and infrared (IR) imagery. However, due to UAV platform motion and asynchronous imaging, spatial misalignment frequently occurs between modalities, leading to weak alignment. This introduces two major challenges: semantic inconsistency at corresponding spatial locations and modality conflict during feature fusion. Existing methods often address these issues in isolation, limiting their effectiveness. In this paper, we propose Cross-modal Offset-guided Dynamic Alignment and Fusion (CoDAF), a unified framework that jointly tackles both challenges in weakly aligned UAV-based object detection. CoDAF comprises two novel modules: the Offset-guided Semantic Alignment (OSA), which estimates attention-based spatial offsets and uses deformable convolution guided by a shared semantic space to align features more precisely; and the Dynamic Attention-guided Fusion Module (DAFM), which adaptively balances modality contributions through gating and refines fused features via spatial–channel dual attention. By integrating alignment and fusion in a unified design, CoDAF enables robust UAV object detection. Experiments on standard benchmarks validate the effectiveness of our approach, with CoDAF achieving a mAP of 78.6\% on the DroneVehicle dataset.

\end{abstract}


\section{Introduction}

\IEEEPARstart{O}{bject} detection from unmanned aerial vehicles (UAVs) is critical for a range of remote sensing tasks, including disaster response~\cite{disaster}, infrastructure inspection~\cite{yolo-mf,zhou2025gca2net}, smart agriculture~\cite{leaf}, and pedestrian detection~\cite{rural}. However, conventional single-modal detectors based on visible (RGB) imagery often suffer in adverse environments such as low illumination or inclement weather. To overcome these limitations, multimodal UAV object detection, particularly approaches~\cite{c2former,calnet,wmaf} that integrate visible and infrared (RGB-IR) imagery, has attracted growing attention due to its ability to exploit complementary spectral information for robust, all-day object detection under diverse environmental conditions~\cite{e2e,ctemrs}.

\begin{figure}[h]
\centering
\subfloat[{\fontsize{7pt}{7pt}\selectfont Semantic inconsistency problem} \label{fig:misaligned}]{
    \includegraphics[width=0.49\linewidth]{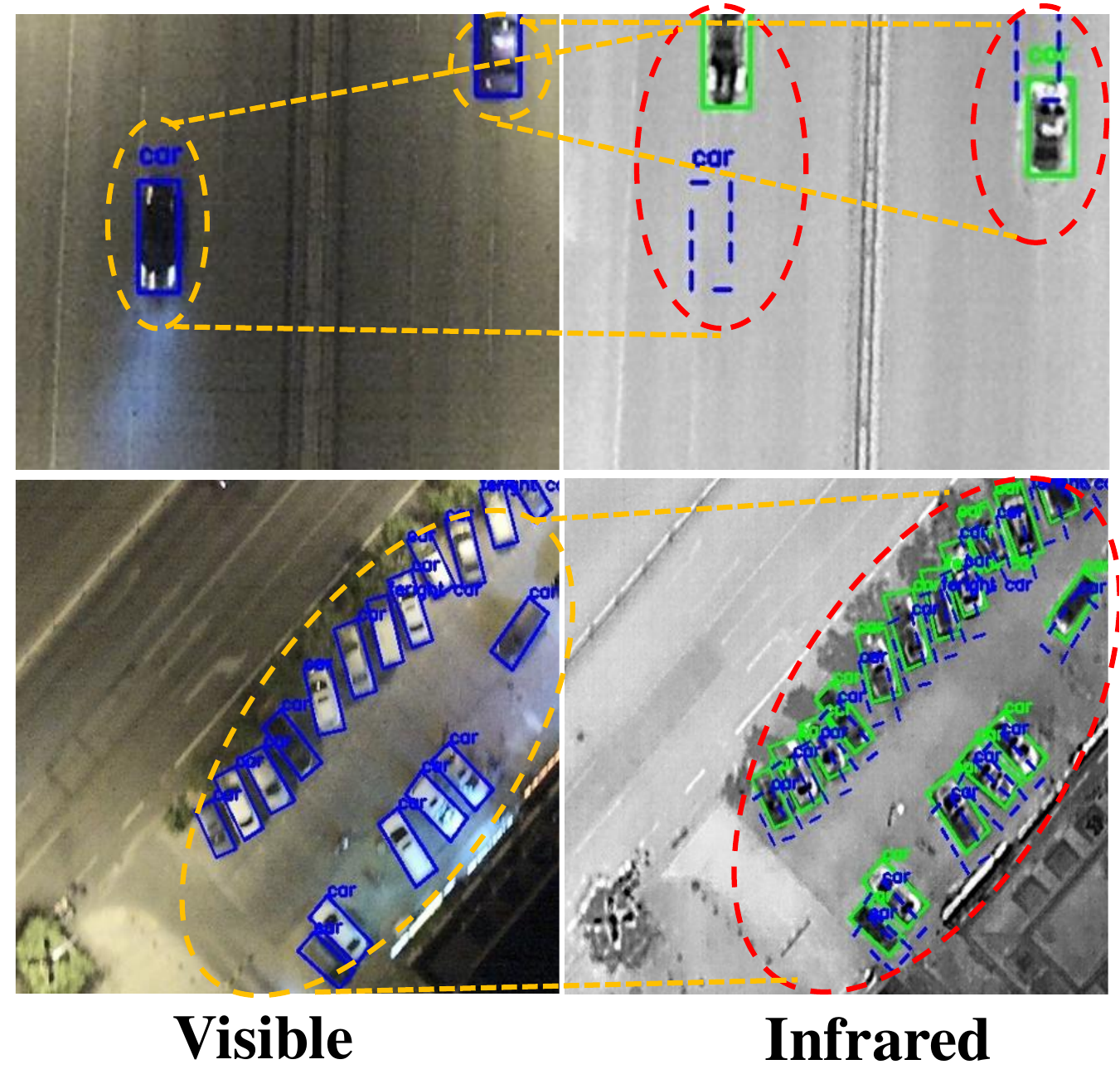}}
\subfloat[{\fontsize{7pt}{7pt}\selectfont Modality conflict problem} \label{fig:conflict}]{
    \includegraphics[width=0.49\linewidth]{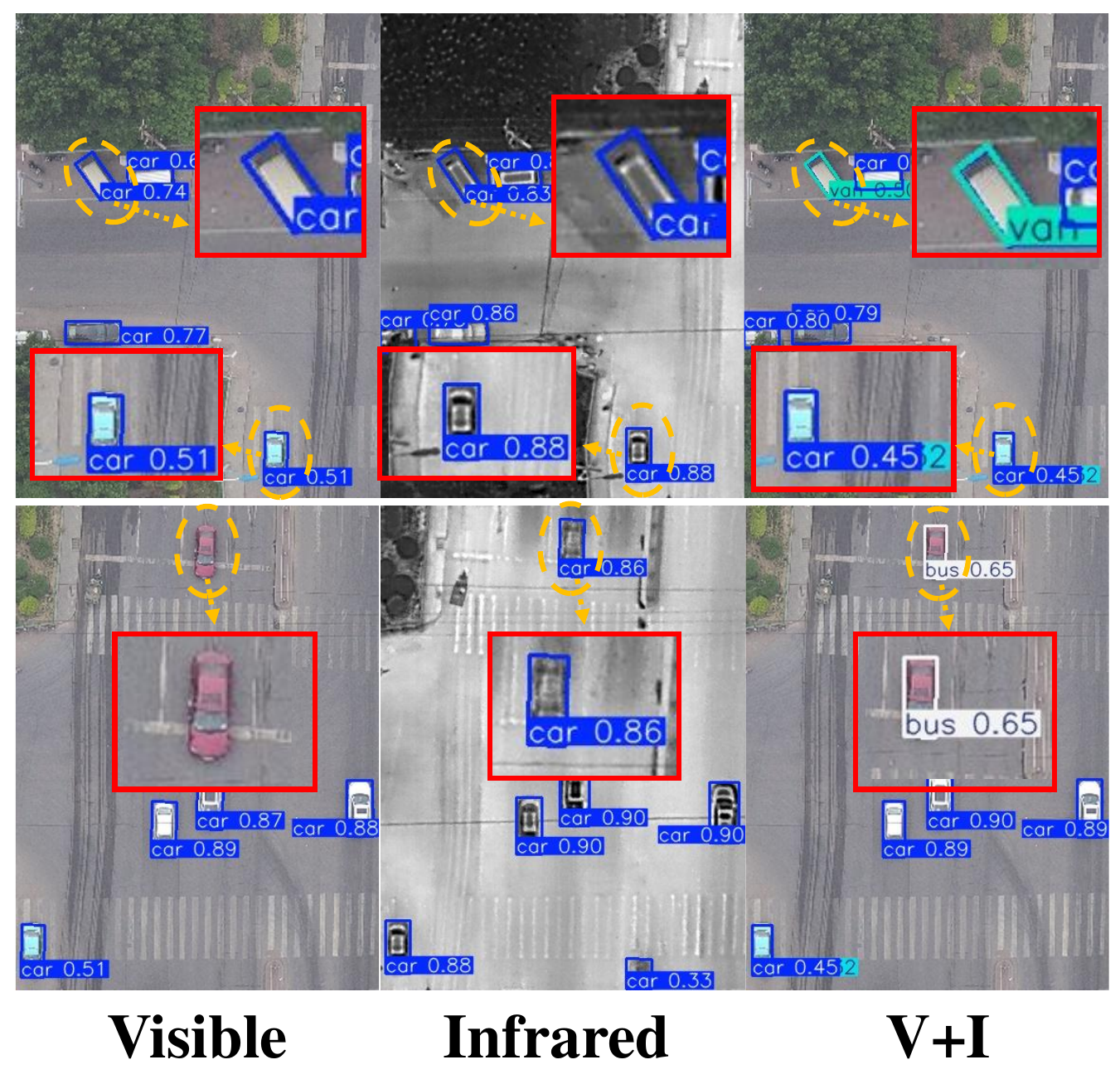}}
\caption{Examples of the semantic inconsistency and the modality conflict problem in RGB-IR UAV object detection. (a) illustrates semantic inconsistency in the DroneVehicle~\cite{uacmdet} dataset, where green and blue bounding boxes denote annotations for the same object in infrared and visible images, respectively. (b) highlights the modality conflict issue, which leads to undetected, incorrectly detection, and conflict detection.}
\label{fig1}
\end{figure}

Despite its advantages, multimodal UAV object detection is highly sensitive to alignment between visible and infrared modalities~\cite{calnet}. Achieving such alignment in UAV-based systems is particularly challenging. In UAV scenarios, platform motion often introduces minor pose variations, causing geometric misalignments between RGB and IR sensors~\cite{uacmdet}. Furthermore, asynchronous imaging due to sensor-level differences can result in spatial shifts of dynamic objects, particularly during high-speed flight. These issues collectively contribute to weak cross-modal alignment, thereby introducing significant obstacles to effective feature fusion and ultimately diminishing detection performance.


The above weak alignment gives rise to two primary challenges for effective multimodal object detection: \textbf{semantic inconsistency} and \textbf{modality conflict}.  From Fig.~\ref{fig1}\subref{fig:misaligned}, semantic inconsistency occurs when corresponding regions across modalities do not carry consistent semantic content due to spatial misalignment, thereby disrupting reliable feature correspondence and impairing cross-modal fusion. In parallel, modality conflict arises when features from different modalities represent contradictory or redundant information, often leading to confusion between objects and background or among adjacent objects. As illustrated in Fig.~\ref{fig1}\subref{fig:conflict}, modality conflicts degrade discriminative fine-grained features, causing confusion between object-background boundaries and inter-object regions.

Recent studies have proposed various strategies to address  the above two challenges. Zhang et al.\cite{arcnn} introduced a region jittering mechanism to achieve local feature-level alignment and alleviate weak cross-modal alignment. However, the absence of explicit semantic supervision limits its ability to ensure consistent semantic correspondence. Chen et al.\cite{probabilistic46} proposed a training-free method based on probabilistic ensembling, which fuses predictions at the output level to enhance robustness under misalignment. While effective to some extent, it lacks cross-modal feature interaction and relies heavily on unimodal detector performance, hindering precise semantic alignment. Zhang et al.~\cite{cian} tackled modality conflict through simple feature concatenation, but without semantic-level guidance, it may produce inconsistent high-level representations. Notably, these methods tend to focus on either semantic alignment or modality conflict, without addressing both simultaneously, which limits their effectiveness in improving detection performance.

To address the above challenges in a unified manner, we propose Cross-modal Offset-guided Dynamic Alignment and Fusion (CoDAF), an adaptive alignment and fusion framework tailored for weakly aligned UAV object detection. 
CoDAF is designed to mitigate semantic inconsistency and modality conflict through two key modules: the Offset-guided Semantic Alignment (OSA) and the Dynamic Attention-guided Fusion Module (DAFM).
OSA aims to resolve semantic inconsistency at corresponding spatial locations across modalities, by estimating offsets from differences in attention responses between infrared and visible features. These offsets guide deformable convolutions for spatial correction, enhancing alignment accuracy. To ensure stable and semantically consistent alignment, OSA further incorporates a Shared feature extraction module (SID) that maps both modalities into a unified semantic space, serving as a reference for more reliable offset prediction and reducing modality-specific noise.
DAFM mitigates modality conflict through a combination of two collaborative sub-modules. The Modality-Adaptive Gating Network (MAGN) adaptively regulates the contribution of each modality based on contextual cues, and its output is further refined by the Dual-Attention Cross-Modulation (DACM), which jointly modulates spatial and channel features to enhance semantic complementarity and suppress conflicts. Extensive experimental results on popular datasets show that our proposed framework significantly improving the performance of weakly aligned UAV object detection.

We highlight our main contributions as follows:
\begin{itemize}
    \item To address semantic inconsistency and modality conflict in weakly aligned UAV-based scenarios, we propose a novel and effective framework named CoDAF, which unifies alignment and fusion in a single design.

    \item To improve cross-modal spatial consistency, we design OSA, which estimates offsets from attention differences and applies deformable convolution for alignment, assisted by SID that projects features into a shared semantic space to guide offset prediction.

    \item To enhance cross-modal feature fusion, we develop DAFM, which combines MAGN to adaptively regulate modality contributions, and DACM to jointly refine features across spatial and channel dimensions for better semantic complementarity.

    \item Extensive experiments demonstrate that our CoDAF framework achieves state-of-the-art performance on cross-moda object detection benchmarks, validating the effectiveness of the proposed modules under weak alignment conditions.
\end{itemize} 

\section{Related Work}

\subsection{Aerial Object Detection}

Oriented object detection focuses on accurately localizing objects with arbitrary orientations, which is essential in aerial imagery, scene text detection, and industrial inspection. Unlike traditional detectors that rely on horizontal bounding boxes, oriented detectors adopt rotated bounding boxes to better capture object shapes and directions. To address the challenges of angle prediction and feature misalignment, early methods such as R²CNN~\cite{r2cnn} and RRPN~\cite{rrpn} introduced rotation-sensitive region proposals with angle regression branches. However, they often suffered from boundary discontinuities and low-quality proposals. RoI Transformer~\cite{roitrans} alleviated these issues by applying a learnable spatial transformer to align region features with rotated RoIs, achieving notable improvements in aerial benchmarks. Based on this, S²A-Net~\cite{s2anet} proposed a dual-branch alignment framework that integrates spatial and semantic alignment to enhance rotated feature extraction. R3Det~\cite{r3det} further addressed feature misalignment through multistage refinement and designed a loss function robust to angle discontinuities, enhancing overall detection robustness. Gliding Vertex~\cite{gv} introduced a vertex sliding mechanism that decouples rotation prediction from bounding-box regression, enabling more accurate localization for densely packed or elongated objects. In parallel, AO2-DETR~\cite{ao2detr} combined transformer-based detection with rotation-sensitive attention and positional encoding, achieving fully end-to-end oriented detection without relying on predefined anchors. To reduce annotation cost, weakly or semisupervised approaches have also emerged. Point-to-RBox~\cite{pointrbox} regresses high-quality rotated boxes using only point-level supervision, while SOOD~\cite{sood} adopts a semi-supervised learning framework combining synthetic data and pseudo-labels to train orientation-aware detectors with minimal human effort. Despite these advancements, all the above methods are based on single-modality, which limits their robustness in complex or degraded environments.

\subsection{Multi-modal Object Detection}

Previous object detection methods~\cite{caaucd30,dalao31,ECFFNet32} have predominantly relied on RGB images, which are inadequate to capture essential details in low light conditions. However, visible-infrared object detection has emerged as a vital means to address this issue by combining distinct visual information come from visible and infrared images and can enhance object detection performance. Based on early fusion strategies, Konig et al.~\cite{fcrpn33}. proposed a fully convolution fusion Region Proposal Network (RPN) which combined infrared and visible image features using concatenation, concluding that mid-level fusion leads to improved performance~\cite{mdnn34}. Several studies~\cite{modcssa35, caff36} have proposed CNN-based attention modules to more effectively exploit the complementary information between the infrared and visible modalities. Furthermore, transformer-based architectures~\cite{lrafnet37, mmfpt38, icafusion} have also been introduced into infrared-visible object detection to capture a more global complementary relationship between infrared and visible images. In addition to direct feature fusion strategies, some approaches~\cite{fusion39, illumination40, crsiod} leverage illumination information as global guidance to weight the contribution of each modality during fusion. Sun et al.~\cite{lfmdet} employed low-rank filtering experts to adaptively suppress redundant modality-specific components, enabling more effective integration of cross-modal features. Meanwhile, knowledge distillation has emerged as a promising technique for bridging the modality gap; some recent works~\cite{efficient41, 3dd42} utilize cross-modality distillation to enhance thermal detectors by leveraging the richer semantics available in visible images. Despite notable improvements over unimodal methods, many approaches often overlook which features are most suitable for fusion and fail to address the modal conflicts arising from inaccurate fusion strategies.

\subsection{Weakly Aligned Multi-modal Object Detection}

In weakly aligned RGB-IR object detection, direct feature fusion typically results in notable performance degradation. One main solution involves global pixel-wise pre-alignment via deformation field prediction~\cite{murf,rdpn,transmatch,dlfm44,dirbsc43,svfnet45}, which enables strict cross-modal registration. However, these methods are often computationally intensive and impede efficient end-to-end optimization. To address this, Zhang et al.~\cite{arcnn} proposed AR-CNN, which jointly optimizes alignment and detection by jittering regions of interest (ROIs), thereby generating aligned ROI features that improve detection performance. Zhou et al.~\cite{mbnet} introduced an illumination-aware alignment module that adaptively selects and aligns complementary features under varying lighting conditions. Chen et al.~\cite{probabilistic46} proposed a probabilistic ensembling strategy that fuses bounding boxes from different modalities for robust detection without explicit alignment. However, its effectiveness relies heavily on the accuracy of unimodal detectors. Although these methods have made notable progress in mitigating misalignment, they remain inadequate for UAV-based RGB-IR scenarios, where severe spatial discrepancies and modality conflicts persist due to platform motion and asynchronous image capture. To tackle these challenges, we propose CoDAF, which explicitly addresses the semantic inconsistency and modality conflict at the corresponding spatial locations caused by weak alignment, thereby enabling more robust multimodal detection.


\section{Proposed Method}

In Section~\ref{overview}, we provide an overview of the proposed CoDAF. In Sections~\ref{OSA} and ~\ref{DAFM} we elaborate on the proposed OSA and DAFM, respectively. Finally, we introduce the loss function in Section ~\ref{HEAD}. 

\begin{figure*}[htbp]
    \centering
    \includegraphics[width=\textwidth]{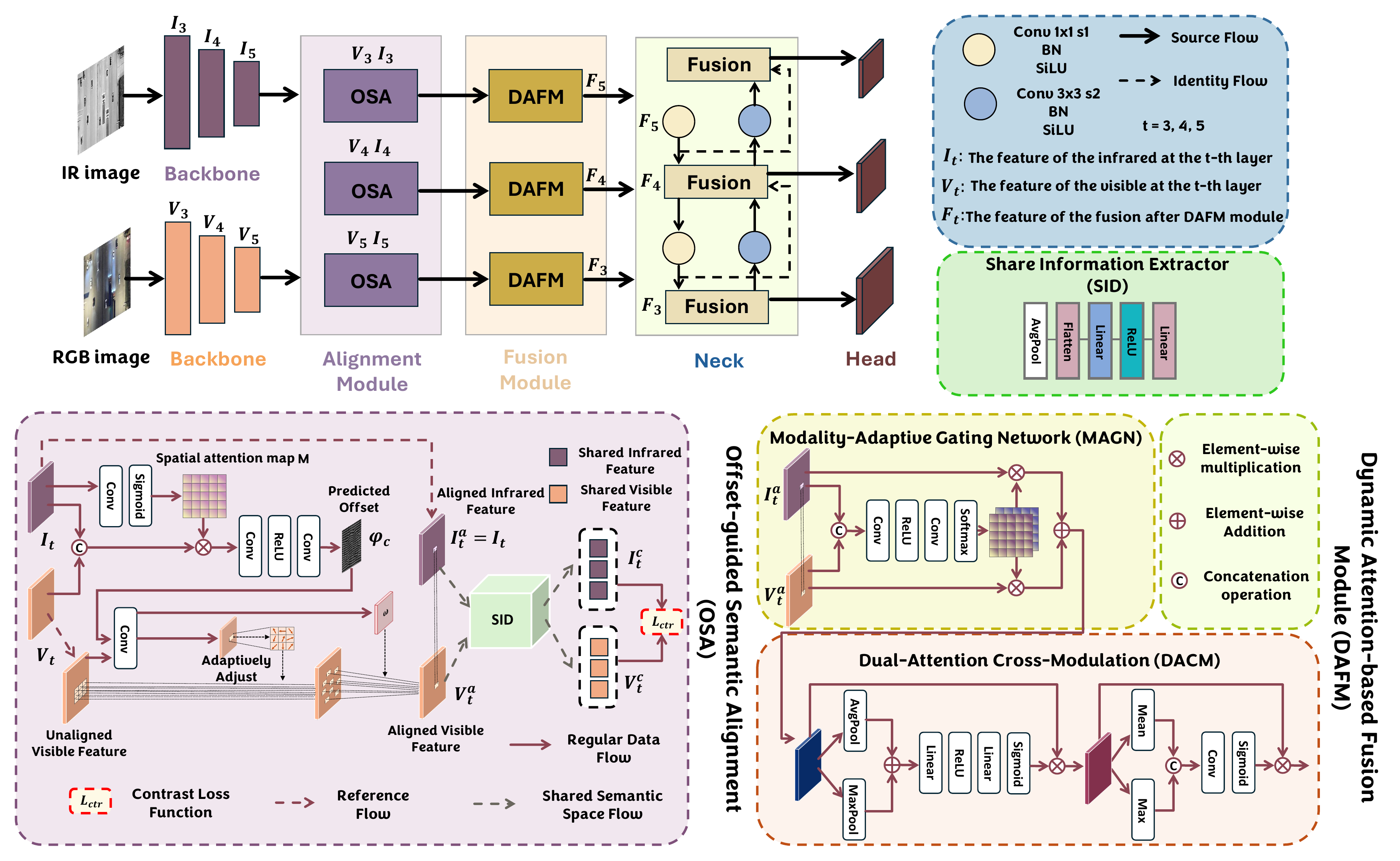} 
    \caption{Overview of the proposed CoDAF framework, which comprises the Offset-guided Semantic Alignment (OSA) and the Dynamic Attention-guided Fusion Module (DAFM). OSA aligns RGB features to the IR domain using attention-guided offsets and deformable convolution, while DAFM adaptively fuses aligned features via modality-aware gating and dual attention. The fused features are forwarded to the neck and detection head to generate the final detection results.}
    \label{fig2}
\end{figure*}

\subsection{Overview}
\label{overview}
The overview of our CoDAF is illustrated in Fig.~\ref{fig2}. The framework comprises a two-stream backbone network, three OSA, three DAFM, a neck structure, and three heads. For simplicity, we omit the details of the neck, following the structure of the efficient hybrid encoder of RT-DETR~\cite{rtdetr}. A pair of weakly misaligned infrared and visible images is fed into the two-stream backbone network to acquire the multiscale RGB-IR representations, respectively. Subsequently, the OSA is designed to align the features of the two modalities in spatial positions and semantic information. Then, we develop a DAFM that selects significant features for cross-modal fusion to further alleviate feature misalignment caused by spatial offset and enhance the discriminative properties of the fused features to address the issue of modality conflict. Finally, we input the cross-modal multiscale fused features into the neck structure and the head to obtain the final result.


\subsection{OSA: Offset-guided Semantic Alignment}
\label{OSA}
The existing RGB-IR objection detection methods directly combine features from different modalities to enhance detection accuracy. However, it overlooks that directly using image feature from one modality is hard to correctly match cross-modality representations in spatial positions under weakly misalignment conditions. To address this issue, we introduce Offset-guided Semantic Alignment (OSA), as shown in Fig.~\ref{fig2}. Given the RGB features $\left\{\mathcal{V}_{t} \in \mathbb{R}^{H \times W \times C}\ \mid t = 3, 4, 5 \right\}$ and the IR features $\left\{\mathcal{I}_{t} \in \mathbb{R}^{H \times W \times C}\ \mid t = 3, 4, 5 \right\}$ obtained from the last three layers of the backbone network, we feed them into the OSA. The feature maps have a size of ${H \times W \times C}$, where $H$ and $W$ are height and width. $C=3$ for RGB images or $C=1$ for IR images. 
To localize semantically relevant regions and improve cross-modal alignment, we first compute a spatial attention map $\mathcal{M}$ from the IR features:
\begin{equation}
\mathcal{M} = \sigma \left(\mathcal{C}_{1 \times 1}(\mathcal{I}_{t}) \right),
\label{eq1}
\end{equation}
where $\mathcal{C}_{1 \times 1}(\cdot)$ denotes a $1 \times 1$ convolution layer and $\sigma(\cdot)$ is the Sigmoid function. Leveraging the robustness of the infrared modality in adverse conditions, this attention map highlights target-relevant areas while suppressing background noise.

We then concatenate the RGB and IR features and apply the attention map $\mathcal{M}$ to the fused feature map, enhancing the focus on foreground objects and mitigating modality-inconsistent interference. The resulting attention-enhanced features $\mathcal{M}_{w}$ are passed through an offset prediction network composed of two stacked $3 \times 3$ convolution layers followed by a ReLU activation. This network captures spatial discrepancies between modalities and predicts pixel-wise offsets $\varphi_{c}$ to guide subsequent alignment. The entire process enables OSA to generate more reliable cross-modal correspondences by focusing on semantically aligned regions. Formally, we define this process as follows:
\begin{equation}
\mathcal{M}_{w} = \mathcal{M} \cdot \mathcal{C}at( \mathcal{I}_{t}, \mathcal{V}_{t}),
\label{eq2}
\end{equation}
\begin{equation}
\varphi_{c} = \mathcal{C}_{3 \times 3}(ReLU(\mathcal{C}_{3 \times 3}(\mathcal{M}_{w}))),
\label{eq3}
\end{equation}
where $\mathcal{C}at(\cdot)$ indicates the channel-wise concatenation operation; $\mathcal{C}_{3 \times 3}(\cdot)$ denotes the operation of a $3 \times 3$ convolution.

Inspired by~\cite{deformabledetr,wmaf}, we leverage a predicted base offset $\varphi_{c}$ to guide the alignment process with stronger spatial priors. To enable implicit refinement and adaptive feature alignment, we integrate deformable convolution-v2 (DCNv2)~\cite{dcnv2} into our framework. Unlike standard DCNv2, which directly learns offsets from input features, our method incorporates $\varphi_{c}$ as a prior to modulate the learned offsets, leading to a more accurate and stable alignment. Specifically, given the central sampling position $x(p)$ in the unaligned feature map $\mathcal{V}_t$ and the predicted base offset $\varphi_c$, the alignment of the feature location $y(p)$ in the refined feature map $\mathcal{V}_{t}^{a}$ is calculated as:
\begin{equation}
y(p) = \sum_{k=1}^{K} w_k \cdot x(p +  \varphi_{c} + p_k + \Delta p_k) \cdot \Delta m_k,
\label{eq4}
\end{equation}
where $K$ denotes the number of sampling points, the modulation scalar $\Delta m_k$ lies in the range $\left[0,1\right]$ to dynamically aggregate information around the corresponding position $p$. The $w_k$ and $p_k$ represent the $k$-th convolution weight and the $k$-th predefined offset, respectively. For example, \( p_k \in \{(-1,1), (-1,0), \dots, (1,1)\} \) defines a regular $3 \times 3$ convolution grid with $K = 9$. The learned residual offset $\Delta p_k$, which represents the relative refinement of the base offset at the $k$-th sampling position, is calculated as $\Delta p_k = \text{Conv}(x)[p + \varphi_c]$, where $\varphi_c$ serves as a strong prior to regularizing the offset learning process. Since the final sampling positions $p + \varphi_c + p_k + \Delta p_k$ may lie in fractional coordinates, bilinear interpolation is applied to obtain the corresponding feature values. By introducing $\varphi_c$ as an explicit spatial prior, our approach not only enhances the precision of cross-modal alignment, but also accelerates convergence during training.

However, due to the inherent differences in feature distributions between infrared and visible modalities, such modality discrepancies can negatively impact the accuracy of offset estimation. To mitigate this issue, we introduce a Shared Feature Extractor module (SID) to construct a shared semantic space for both modalities. Specifically, SID leverages a contrastive learning strategy to extract modality-invariant features shared between infrared and visible images. These features capture consistent semantic cues, such as object structure and scene layout. As the infrared modality serves as the alignment reference, we denote the aligned infrared feature by $\mathcal{I}_{t}^{a} = \mathcal{I}_{t}$, while $\mathcal{V}_{t}^{a}$ is adaptively aligned to it. The extracted shared representations are further employed as a distribution alignment mapping to assess spatial consistency across modalities, thereby facilitating a more accurate offset estimation. The SID module $f_{SID}$ consists of two linear layers, a ReLU activation layer, a flatten layer, a normalization layer, and an AdaptiveAvgPool layer. The modality-invariant features $\mathcal{V}_{t}^{c}$ and $\mathcal{I}_{t}^{c}$ are obtained as:
\begin{equation}    
\mathcal{V}_{t}^{c} = Norm(f_{SID}(\mathcal{V}_{t}^{a})),
\label{eq5}
\end{equation}
\begin{equation}
\mathcal{I}_{t}^{c} = Norm(f_{SID}(\mathcal{I}_{t}^{a})).
\label{eq6}
\end{equation}

The contrastive loss function $\mathcal{L}_{contrast}$ is formalized based on the InfoNCE~\cite{infonce} objective as follows:
\begin{equation}
\mathcal{L}_{contrast} = -\frac{1}{N} \sum_{i=1}^{N} \log \left( \frac{\exp\left(\left({\mathcal{V}_{t,i}^{c} \cdot \mathcal{I}_{t,i}^{c}}\right)/{\tau}\right)}{\sum_{j=1}^{N} \exp\left(\left({\mathcal{V}_{t,i}^{c} \cdot \mathcal{I}_{t,j}^{c}}\right)/{\tau}\right)} \right),
\label{eq7}
\end{equation}
where $N$ denotes the total number of training samples in the batch and $\tau$ is a temperature hyperparameter that adjusts the sharpness of the similarity distribution. For each $\mathcal{V}_{t,i}^{c}$, the positive sample is its paired IR feature $\mathcal{I}_{t,i}^{c}$, while all $N$ candidates $\{ \mathcal{I}_{t,j}^{c} \}_{j=1}^{N}$ in the batch are used to compute the softmax denominator. This contrastive objective encourages high similarity between paired visible and infrared features while suppressing similarity with unpaired samples, thereby promoting consistent cross-modal alignment.

We apply two types of constraints for spatial alignment: spatial alignment loss $\mathcal{L}_{sm}$ and attention loss $\mathcal{L}_{attn}$. The spatial alignment loss is made up of two components: the structural similarity index measure~\cite{ssim} (SSIM) $\mathcal{L}_{ssim}$, and the L1 losses $\mathcal{L}_{mae}$. The alignment loss is denoted as: 
\begin{equation}
\begin{aligned}
\mathcal{L}_{sm} = \lambda_{1} \mathcal{L}_{ssim} (\mathcal{V}_{t}^{a},\mathcal{I}_{t}^{a}) + \lambda_{2} \mathcal{L}_{mae} (\mathcal{V}_{t}^{a},\mathcal{I}_{t}^{a}),
\label{eq8}
\end{aligned}
\end{equation}
where we set $\lambda_{1}$ and $\lambda_{2}$ to 0.3 and 0.5 respectively. These components ensure multi-dimensional consistency between the features aligned by deformable convolutions and the target modality, especially in terms of structure and pixel values.

To effectively guide the attention map $\mathcal{M}$ towards target regions while ensuring spatial coherence, we introduce the attention loss $\mathcal{L}_{attn}$, which consists of two regularization terms. The sparsity loss $\mathcal{L}_{sparse}$ encourages compact activation in informative regions by penalizing widespread responses and is defined as:
\begin{equation}
\begin{aligned}
\mathcal{L}_{sparse} = -\frac{1}{B H W} \sum_{b=1}^B \sum_{h=1}^H \sum_{w=1}^W \mathcal{M}_{b,1,h,w} \log\left(\mathcal{M}_{b,1,h,w} + \epsilon \right),
\label{eq9}
\end{aligned}
\end{equation}
while smoothness loss $\mathcal{L}_{smooth}$ enforces spatial continuity by penalizing abrupt variations in horizontal and vertical directions:
\begin{equation}
\begin{aligned}
\mathcal{L}_{smooth} = \frac{1}{B H (W-1)} \sum_{b=1}^B \sum_{h=1}^H \sum_{w=1}^{W-1} \left| \mathcal{M}_{b,1,h,w+1} - \mathcal{M}_{b,1,h,w} \right| \\
+ \frac{1}{B (H-1) W} \sum_{b=1}^B \sum_{h=1}^{H-1} \sum_{w=1}^W \left| \mathcal{M}_{b,1,h+1,w} - \mathcal{M}_{b,1,h,w} \right|.
\label{eq10}
\end{aligned}
\end{equation}
The formula~\ref{eq9} and formula~\ref{eq10} are combined to form the overall attention loss:
\begin{equation}
\begin{aligned}
\mathcal{L}_{attn} = \mathcal{L}_{sparse} + \lambda_{3}\mathcal{L}_{smooth},
\label{eq11}
\end{aligned}
\end{equation}
where $\epsilon$ and $\lambda_{3}$ are hyper-parameters set to $10^{-8}$ and $0.1$, respectively. The $\mathcal{L}_{attn}$ is jointly optimized with the spatial alignment loss $\mathcal{L}_{sm}$ and the contrastive loss $\mathcal{L}_{contrast}$ to form the total alignment loss:
\begin{equation}
\begin{aligned}
\mathcal{L}_{t} = \mathcal{L}_{contrast} + \mathcal{L}_{sm} + \mathcal{L}_{attn},
\label{eq12}
\end{aligned}
\end{equation}
Through iterative joint optimization under the strategy of \textbf{`alignment first, then comparison'}, the model progressively refines spatial and semantic alignment from coarse to fine, enhancing robustness in challenging sensing conditions.

\subsection{DAFM: Dynamic Attention-based Fusion Module}
\label{DAFM}
Prior approaches often utilize fixed parameters to balance the contributions of RGB and IR modalities in feature fusion. However, static weighting fails to address variations in modality reliability that lead to the issue of modal conflicts. For instance, a region with high attention in the infrared modality may not have corresponding visible modality (\textit{e.g.}, under low light conditions where RGB images have limited information). After feature fusion, the fused feature map actually pays less attention to that region, leading to a waste of the infrared modality's features that leads to object classification confusion and mislocation, impacting the final detection results. 

To overcome this limitation, we propose DAFM, which comprises two components: a Modality-Adaptive Gating Network (MAGN) and a Dual-Attention Cross-Modulation (DACM), as illustrated in Fig.~\ref{fig2}. DAFM adaptively modulates the contribution of each modality at the spatial level and exploits cross-modal interactions to enhance the discriminability of the fused features. MAGN dynamically assigns pixel-wise importance weights to RGB and IR features by jointly analyzing their semantic representations. Given the aligned RGB feature map $\mathcal{V}_{t}^{a}$ and the IR feature map $\mathcal{I}_{t}^{a}$, the gating and fusion process is formulated as:
\begin{equation}
\begin{aligned}
\mathcal{G} = \text{Softmax} \left( \mathcal{C}_{3 \times 3} \left( ReLU \left(\mathcal{C}_{3 \times 3} \left(\mathcal{C}at \left(\mathcal{V}_{t}^{a}, \mathcal{I}_{t}^{a}\right)  \right)\right)\right) \right),
\label{eq13}
\end{aligned}
\end{equation}
\begin{equation}
\begin{aligned}
\mathcal{F} = \mathcal{G}_{v} \odot \mathcal{V}_{t}^{a} + \mathcal{G}_{i} \odot \mathcal{I}_{t}^{a},
\label{eq14}
\end{aligned}
\end{equation}
where $\odot$ denotes element-wise multiplication, and $\mathcal{G} \in \mathbb{R}^{2 \times H \times W}$ represents the spatially adaptive weighting map that assigns pixel-wise importance to each modality. $\mathcal{G}_{v}$ and $\mathcal{G}_{i}$ are modality-specific weighting maps for RGB and IR, respectively, derived from $\mathcal{G}$ along the modality channel. These weights are used to selectively emphasize more informative modality cues at each spatial location, resulting in the fused feature map $\mathcal{F}$. This gating mechanism enables the integration of adaptive features, enhancing the robustness of the network under diverse visual conditions.

To suppress modality-specific noise and enhance semantic complementarity, we introduce DACM, which hierarchically applies channel and spatial attention to refine the fused features $\mathcal{F}$. Channel attention first captures inter-channel dependencies by aggregating global context via average and max pooling, whose sum is passed through a shared MLP and sigmoid activation to produce the attention map $\mathcal{H}_c$. The channel-refined feature $\mathcal{F}_c$ is obtained by element-wise multiplication with $\mathcal{F}$, as defined by:
\begin{equation}
\begin{aligned}
\mathcal{H}_c = \sigma \left( \text{MLP}\left(\text{AvgPool}(\mathcal{F}) + \text{MaxPool}(\mathcal{F}) \right)\right),
\label{eq15}
\end{aligned}
\end{equation}
\begin{equation}
\begin{aligned}
\mathcal{F}_c = \mathcal{H}_c \odot \mathcal{F},
\label{eq16}
\end{aligned}
\end{equation}
where $\text{MLP}(\cdot)$ denotes a lightweight bottleneck consisting of two linear layers with ReLU activation, $\text{AvgPool}(\cdot)$ and $\text{MaxPool}(\cdot)$ represent the average pooling operation and the maximum pooling operation, respectively.

Subsequently, spatial attention is applied to highlight modality-consistent and spatially salient regions. A spatial attention map $\mathcal{H}_s$ is computed by concatenating the channel-wise average and max pooled features of $\mathcal{F}_c$, followed by a $3 \times 3$ convolution and a sigmoid activation. The final fused representation $\mathcal{F}_{\text{fused}}$ is obtained by modulating $\mathcal{F}_c$ with $\mathcal{H}_s$, as formulated below:
\begin{equation}
\begin{aligned}
\mathcal{H}_s = \sigma \left( \mathcal{C}_{3 \times 3} \left( \mathcal{C}at(\text{Mean}(\mathcal{F}_c), \text{Max}(\mathcal{F}_c))  \right) \right),
\label{eq17}
\end{aligned}
\end{equation}
\begin{equation}
\begin{aligned}
\mathcal{F}_{\text{fused}} = \mathcal{H}_s \odot \mathcal{F}_c,
\label{eq18}
\end{aligned}
\end{equation}
where $\text{Mean}(\cdot)$ and $\text{Max}(\cdot)$ denote spatial pooling operations. This hierarchical attention design adaptively filters out irrelevant modality-specific noise while enhancing spatially aligned semantics, thereby improving robustness under challenging conditions such as low light or clutter.

\subsection{Loss Function}
\label{HEAD}

Our overall approach is built upon a supervised detection framework, where the objective of multimodal alignment is not to achieve strict pixel-wise correspondence, but to facilitate optimal feature-level adaptation that enhances detection performance. To this end, the alignment mechanism is seamlessly integrated into the detection pipeline rather than optimized as an independent objective. The overall training loss of our CoDAF framework is defined as:
\begin{equation}
\begin{aligned}
\mathcal{L}_{\text{total}} = \mathcal{L}_{\text{det}} +  \lambda \mathcal{L}_{t},
\label{eq19}
\end{aligned}
\end{equation}
where $\lambda$ is a balancing hyperparameter, empirically set to 0.1 in our experiments. For the oriented bounding box (OBB) detection task, we adopt the detection loss $\mathcal{L}_{\text{det}}$ and OBB head from YOLOv11-OBB~\cite{yolov11}. On the $M3FD$~\cite{tardal} dataset, we utilize the detection loss and prediction head consistent with those of RT-DETR.

\section{Experiments}
In this section, we first describe the implementation details of our approach. Then, we conduct a comprehensive comparison between the proposed CoDAF and several state-of-the-art methods. Finally, we perform ablation studies to evaluate the contribution and effectiveness of each key component within our framework.
\subsection{Datasets and Metrics}
We conduct experiments on two publicly available datasets, encompassing objects with varying scales, modalities, and scene complexities. To quantitatively evaluate detection performance, we adopt the standard COCO Average Precision (mAP@.5 and mAP@.5:.95) metric. The datasets used in our experiments include DroneVehicle~\cite{uacmdet} and $M3FD$~\cite{tardal}.
\subsubsection{\textbf{DroneVehicle dataset}}
The DroneVehicle dataset serves as the largest and most comprehensive benchmark for cross-modal vehicle detection in UAV-based imagery. It contains 28,439 pairs of visible and infrared images, with a total of 953,087 manually annotated vehicle instances categorized into five classes: car, bus, truck, van, and freight car. These image pairs were captured by UAV under a wide range of viewing angles, flight altitudes, illumination conditions, and diverse real-world environments, including urban streets, residential zones, parking areas, and highways, covering both daytime and nighttime scenes. The dataset is partitioned into 17,990 pairs for training, 1,469 pairs for validation, and 8,980 pairs for testing. Each image has a spatial resolution of $840 \times 712$ pixels, obtained by adding a 100-pixel white margin to the original $640 \times 512$ image to preserve alignment boundaries.

Although the dataset incorporates pre-registration processes to spatially align RGB-IR image pairs, weak misalignment problems remain. These arise mainly from the time discrepancies in image acquisition and the dynamic movement of targets between the two modalities, resulting in local spatial inconsistencies across corresponding image regions. Approximately 35\% of the bounding boxes in the DroneVehicle dataset exhibit noticeable misalignment, with most deviation distances falling within the range of 0 to 15 pixels~\cite{wmaf}.

\subsubsection{\textbf{\textit{M3FD} dataset}}
The $M3FD$ dataset comprises 4,200 pairs of infrared and visible images with a spatial resolution of $1024 \times 768$ pixels, covering a wide range of scenes and six predefined object categories. While the dataset exhibits slight spatial misalignment between image pairs, it provides valuable multi-modal information for object detection under diverse conditions. As the dataset does not include an official split, we partition the data into 3,368 image pairs for training and 831 pairs for validation, based on different scene distributions. This manual split introduces low scene similarity between the training and validation subsets, thereby increasing the difficulty of the cross-modal detection task and providing a more rigorous evaluation of model generalization across varying environments.

\subsection{Implementation Details}
We implement our CoDAF framework using the PyTorch library on a system equipped with 8 RTX 3090 GPUs. ResNet-50 is adopted as the backbone,  initialized with pre-trained weights from the COCO dataset. The model is optimized using the AdamW optimizer with a batch size of 48, a learning rate of 0.0001, and a weight decay of 0.0001. To ensure fair comparisons with state-of-the-art approaches, input images are resized to $640 \times 640$ for both training and testing on the DroneVehicle and $M3FD$ datasets. The model is trained for 75 epochs on DroneVehicle and 150 epochs on $M3FD$. Notably, for the DroneVehicle dataset, the ground truth annotations from the IR images are used as training labels, as they provide more comprehensive and accurate target annotations compared to the visible modality. Notably, we remove both the proposed OSA and DAFM, fusing the features through simple element-wise addition to form the \textbf{Baseline} model.

\begin{figure*}[htbp]
    \centering
    \includegraphics[width=\textwidth]{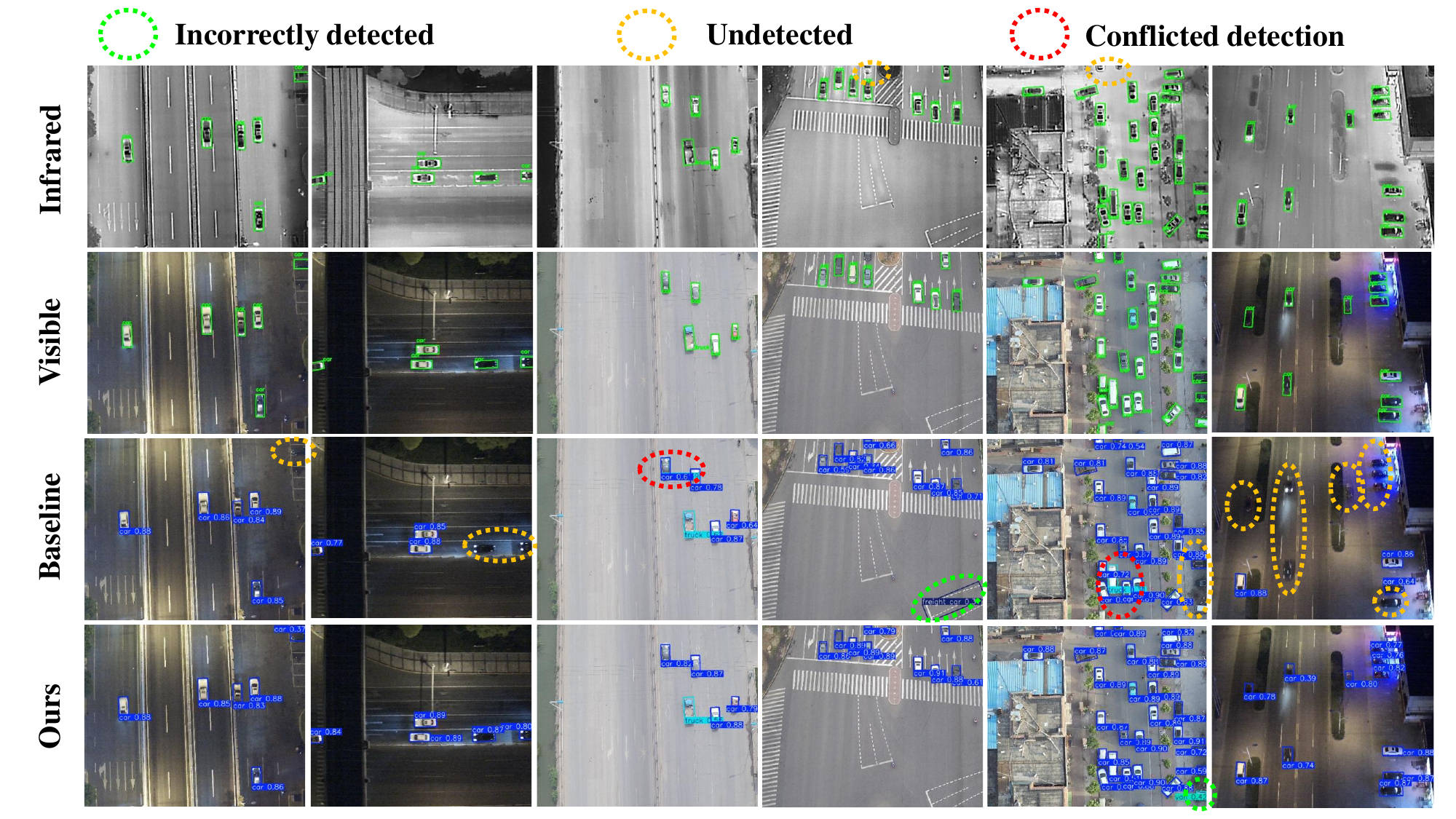} 
    \caption{Visualization results of CoDAF on the DroneVehicle dataset under weakly misaligned conditions. Different object categories are annotated with color-coded bounding boxes for clear differentiation. The model demonstrates strong performance in detecting small-scale objects, handling densely packed scenes, and achieving fine-grained classification across diverse lighting scenarios. Detection errors are highlighted using dashed circles: \textcolor{green}{green} for false positives, \textcolor{orange}{orange} for missed detections, and \textcolor{red}{red} for conflicting predictions.}
    \label{fig3}
\end{figure*}

\begin{figure}[htbp]
    \centering
    \includegraphics[width=\columnwidth]{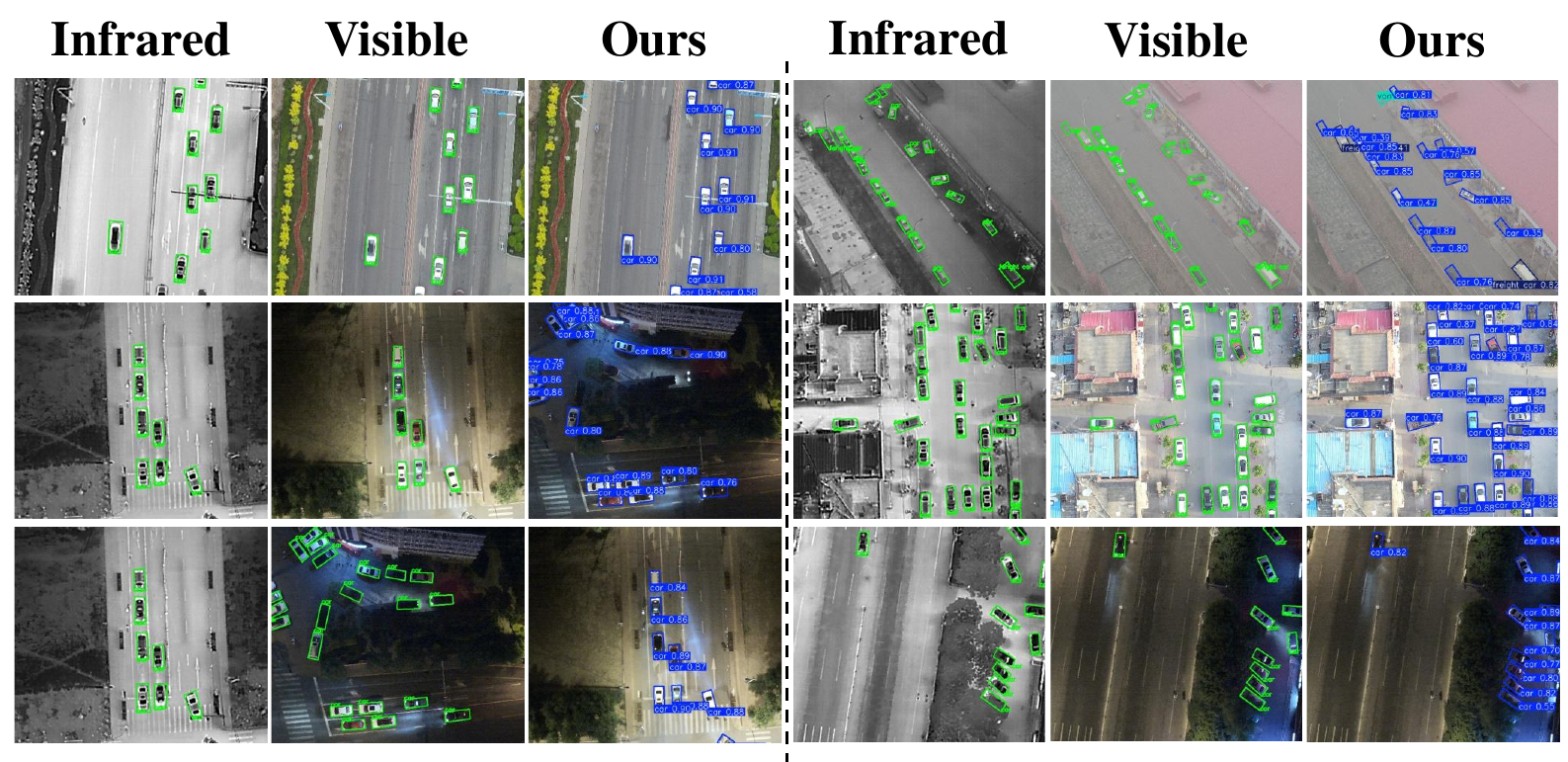} 
    \caption{Detection results of our CoDAF under various challenging conditions on the DroneVehicle dataset, including low-light environments, foggy weather, and daytime scenes with complex backgrounds.}
    \label{fig4}
\end{figure}

\subsection{Comparison Results on DroneVehicle Dataset}

\begin{table*}[ht]
\centering
\caption{Comparisons With State-of-the-Art Methods on the DroneVehicle Dataset. \textbf{Bold}/\underline{Underline} Indicate the Best/Second-Best Results. Note that all detectors locate and classify vehicles with OBB heads. The terms `Vi' and `Ir' refer to the visible and infrared modalities, respectively.}
\renewcommand{\arraystretch}{1.1}
\begin{tabular}{c|c|cc|ccccc|c}
\toprule
\text{Method} & \text{Publication} & \text{Vi} & \text{Ir} & \text{Car} & \text{Truck} & \text{Freight-car} & \text{Bus} & \text{Van} & \text{mAP@.5} \\
\midrule
RetinaNet~\cite{retinanet} & [ICCV '17] & \checkmark & $\times$ & 78.5 & 34.4 & 24.1 & 69.8 & 28.8 & 47.1 \\
Faster R-CNN~\cite{faster} & [NeurIPS '15] & \checkmark & $\times$ & 79.0 & 49.0 & 37.2 & 77.0 & 37.0 & 55.9 \\
ROITrans~\cite{roitrans} & [CVPR '19] & \checkmark & $\times$ & 61.6 & 55.1 & 42.3 & 85.5 & 44.8 & 61.6 \\
YOLOv5s~\cite{yolov5s} & - & \checkmark & $\times$ & 78.6 & 55.3 & 43.8 & 87.1 & 46.0 & 62.1 \\
Oriented R-CNN~\cite{orientrcnn} & [ICCV '21] & \checkmark & $\times$ & 80.1 & 53.8 & 41.6 & 85.4 & 43.3 & 60.8 \\
ReDet~\cite{redet} & [CVPR '21] & \checkmark & $\times$ & 80.3 & 56.1 & 42.7 & 80.2 & 44.4 & 60.8 \\
R$^{3}$Det~\cite{r3det} & [AAAI '21] & \checkmark & $\times$ & 79.3 & 42.2 & 24.5 & 76.0 & 28.5 & 50.1 \\
S$^{2}$ANet~\cite{s2anet} & [TGRS '22] & \checkmark & $\times$ & 80.0 & 54.2 & 42.2 & 84.9 & 43.8 & 61.0 \\
\midrule
RetinaNet~\cite{retinanet} & [ICCV '17] & $\times$ & \checkmark & 88.8 & 35.4 & 39.5 & 76.5 & 32.1 & 54.5 \\
Faster R-CNN~\cite{faster} & [NeurIPS '15] & $\times$ & \checkmark & 89.4 & 53.5 & 48.3 & 87.0 & 42.6 & 64.2 \\
ROITrans~\cite{roitrans} & [CVPR '19] & $\times$ & \checkmark & 90.1 & 60.4 & 58.9 & 89.7 & 52.2 & 70.3 \\
YOLOv5s~\cite{yolov5s} & - & $\times$ & \checkmark & 90.0 & 59.5 & 60.8 & 89.5 & 53.8 & 70.7 \\
Oriented R-CNN~\cite{orientrcnn} & [ICCV '21] & $\times$ & \checkmark & 89.8 & 57.4 & 53.1 & 89.5 & 53.8 & 67.5 \\
ReDet~\cite{redet} & [CVPR '21] & $\times$ & \checkmark & 90.0 & 61.5 & 55.6 & 89.5 & 46.6 & 68.6 \\
R$^{3}$Det~\cite{r3det} & [AAAI '21] & $\times$ & \checkmark & 89.5 & 48.3 & 16.6 & 87.1 & 39.9 & 62.3 \\
S$^{2}$ANet~\cite{s2anet} & [TGRS '22] & $\times$ & \checkmark & 89.9 & 54.5 & 55.8 & 88.9 & 48.4 & 67.5 \\
\midrule
CIAN~\cite{cian} & [IF '19] & \checkmark & \checkmark & 90.1 & 63.8 & 60.7 & 89.1 & 50.3 & 70.8 \\
AR-CNN~\cite{arcnn} & [TNNLS '21] & \checkmark & \checkmark & 90.1 & 64.8 & 62.1 & 89.4 & 51.5 & 71.6 \\
UA-CMDet~\cite{uacmdet} & [TCSVT '22] & \checkmark & \checkmark & 87.5 & 60.7 & 46.8 & 87.1 & 38.0 & 64.0 \\
MBNet~\cite{mbnet} & [ECCV '20]& \checkmark & \checkmark & 90.1 & 64.4 & 62.4 & 88.8 & 53.6 & 71.9 \\
TSFADet~\cite{tsfadet} & [ECCV '22] & \checkmark & \checkmark & 89.9 & 67.9 & 63.7 & 89.8 & 54.0 & 73.1 \\
CALNet~\cite{calnet} & [ACMM '23]& \checkmark & \checkmark & 90.1 & \underline{70.6} & \textbf{67.4} & 89.7 & 55.0 & \underline{74.6} \\
C$^{2}$Former~\cite{c2former} & [TGRS '24] & \checkmark & \checkmark & 90.2 & 68.3 & 64.4 & 89.8 & \underline{58.5} & 74.2 \\
CAGTDet~\cite{cagtdet} & [IF '24] & \checkmark & \checkmark & \underline{90.8} & 69.6 & \underline{66.2} & 90.4 & 55.6  & 74.5 \\
\midrule
\rowcolor[HTML]{b8d6e0} Ours & - & \checkmark & \checkmark & \textbf{93.5} & \textbf{77.9} & \underline{66.2} &  \textbf{90.5} & \textbf{65.0} & \textbf{78.6} \\
\bottomrule
\end{tabular}
\label{tab1}
\end{table*}

\begin{figure}[htbp]
    \centering
    \includegraphics[width=\columnwidth]{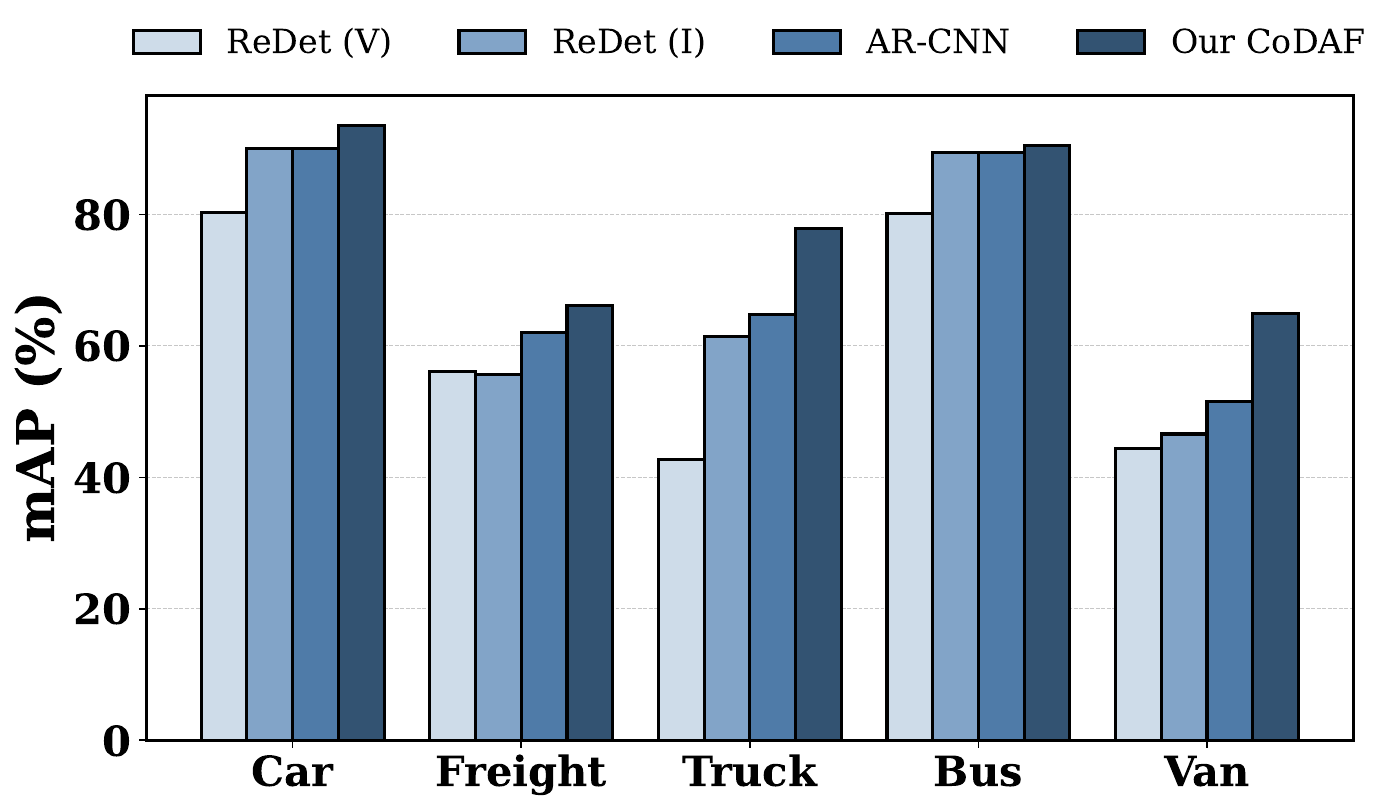} 
    \caption{Comparison of different vehicle detectors on the DroneVehicle dataset. Our CoDAF achieves higher mAP results through a more effective strategy for cross-modality fusion.}
    \label{fig5}
\end{figure}

In this section, We compare our method with state-of-the-art single-modal and multimodal object detection approaches. For single-modal detection, we evaluate several oriented object detectors, including one-stage methods (RetinaNet~\cite{retinanet}, YOLOv5s~\cite{yolov5s}, R$^{3}$Det~\cite{r3det}, S$^{2}$ANet~\cite{s2anet}) and two-stage methods (Faster R-CNN~\cite{faster}, RoITransformer~\cite{roitrans}, Oriented R-CNN~\cite{orientrcnn}, ReDet~\cite{redet}). Each model is trained and tested individually on RGB and IR images to assess their performance on different modalities. For multimodal detection, we compare with eight recent fusion-based methods: CIAN~\cite{cian}, AR-CNN~\cite{arcnn}, UA-CMDet~\cite{uacmdet}, MBNet~\cite{mbnet}, TSFADet~\cite{tsfadet}, CALNet~\cite{calnet}, C$^{2}$Former~\cite{c2former}, and CAGTDet~\cite{cagtdet}. Notably, UA-CMDet, MBNet, TSFADet, C$^{2}$Former, and CAGTDet explicitly address the weak misalignment issue during fusion, highlighting the significance and relevance of this challenge. All methods are implemented using their official settings to ensure a fair comparison.

\subsubsection{\textbf{Quantitative Analysis}}
Table~\ref{tab1} and Fig.~\ref{fig5} show the detection performance of various methods on the DroneVehicle dataset. Compared with unimodal methods, multimodal approaches consistently exhibit superior detection accuracy, highlighting the benefit of cross-modal feature integration. Specifically, methods using only the visible modality suffer from limited robustness in low-light scenes, as the DroneVehicle dataset contains a substantial number of nighttime images. In contrast, infrared-based detectors achieve notably better performance under the same network backbone, demonstrating the advantage of infrared imagery in enhancing feature discriminability under poor illumination. Benefiting from both modalities, CoDAF achieves the highest mAP@.5 of 78.6\%, surpassing the second-best method by 4.0\%, as shown in Table~\ref{tab1}. In particular, CoDAF demonstrates strong discriminative capability in categories prone to spatial and semantic confusion (e.g., \textit{truck}, \textit{freight-car}, and \textit{van}), as illustrated in Fig.~\ref{fig5}. We attribute these improvements to the proposed OSA and DAFM. OSA enables accurate spatial alignment by leveraging modality-aware offsets to correct misaligned features. In parallel, DAFM adaptively modulates cross-modal features via dual attention, enhancing fusion robustness and semantic consistency. Despite achieving superior overall performance, CoDAF shows slightly lower accuracy than CALNet in the \textit{freight-car} category. The reduced performance may be due to CoDAF’s higher sensitivity to spatial scale variations in remote targets, making localization more difficult under weak alignment. Overall, Table~\ref{tab1} substantiate that the proposed alignment-fusion strategy successfully alleviates spatial misalignment and modality interference, enabling more accurate and reliable multimodal detection in challenging UAV environments.

\subsubsection{\textbf{Qualitative Analysis}}

To further validate the effectiveness of CoDAF under weakly aligned multimodal conditions, Fig.~\ref{fig3} presents qualitative comparisons on the DroneVehicle dataset. The first and second rows show ground-truth annotations of infrared and visible modalities, serving as modality-specific references. The third and fourth rows illustrate detection results from the baseline model and CoDAF, respectively. As depicted in Fig.~\ref{fig3}, baseline models often produce false positives, missed detections, and conflicting predictions in low-light and crowded scenes when target positions shift. The main cause lies in the inability to learn properly aligned features under weak alignment, which leads to reduced detection performance. In contrast, CoDAF achieves more robust and accurate detection by extracting discriminative and well-aligned features. The proposed OSA and DAFM effectively alleviate semantic inconsistency and modality conflict caused by weak alignment, leading to more precise and stable multimodal object detection.

\subsubsection{\textbf{Detection Results in Complex Scenarios}}
To demonstrate the effectiveness of our model in challenging conditions, we visualize detection results across various scenarios. As depicted in Fig.~\ref{fig4}, the selected examples include low light, fog, dense object, and daytime. CoDAF consistently delivers accurate detections under these adverse conditions, highlighting its robustness and strong generalization capabilities.

\subsection{Comparison Results on $M3FD$ Dataset}
To verify the effectiveness of our method, we evaluate it on the challenging $M3FD$ dataset, comparing against both unimodal method and a series of state-of-the-art fusion-based detection approaches. Table{~\ref{tab2}} and Fig.~\ref{fig7} present the quantitative comparison results of different unimodal and multimodal detection methods on the $M3FD$ dataset. For unimodal detection, YOLOv5s{~\cite{yolov5s}} is evaluated with either visible or infrared modality. As expected, unimodal method exhibit limited performance due to the modality-specific constraints. For multimodal detection, several existing fusion-based approaches are compared, including DIDFusion{~\cite{didfuse}}, U2Fusion{~\cite{u2fusion}}, PIAFusion{~\cite{piafusion}}, SwinFusion{~\cite{swinfusion}}, Tardal{~\cite{tardal}}, MetaFusion{~\cite{metafusion}}, CDDFuse{~\cite{cddfuse}}, CFT~\cite{cft}, ICAFusion~\cite{icafusion}, and E2E-MFD{~\cite{e2e}}. Most multimodal fusion methods demonstrate notable improvements, confirming the advantage of leveraging cross-modal information from visible and infrared data. Compared with these methods, our approach achieves the best overall performance, reaching 61.2\% mAP@.5:.95 and 90.8\% mAP@.5. This surpasses strong competitors such as CFT~\cite{cft} (60.7\%), E2E-MFD~\cite{e2e} (58.1\%), and MetaFusion~\cite{metafusion} (56.8\%). In particular, our method attains the top results on several challenging categories, including \textbf{Motorcycle} (54.2\%), \textbf{Lamp} (50.7\%), and ranks second in \textbf{Truck} (68.4\%), indicating superior robustness to small and low-contrast targets.

\begin{figure*}[htbp]
    \centering
    \includegraphics[width=\textwidth]{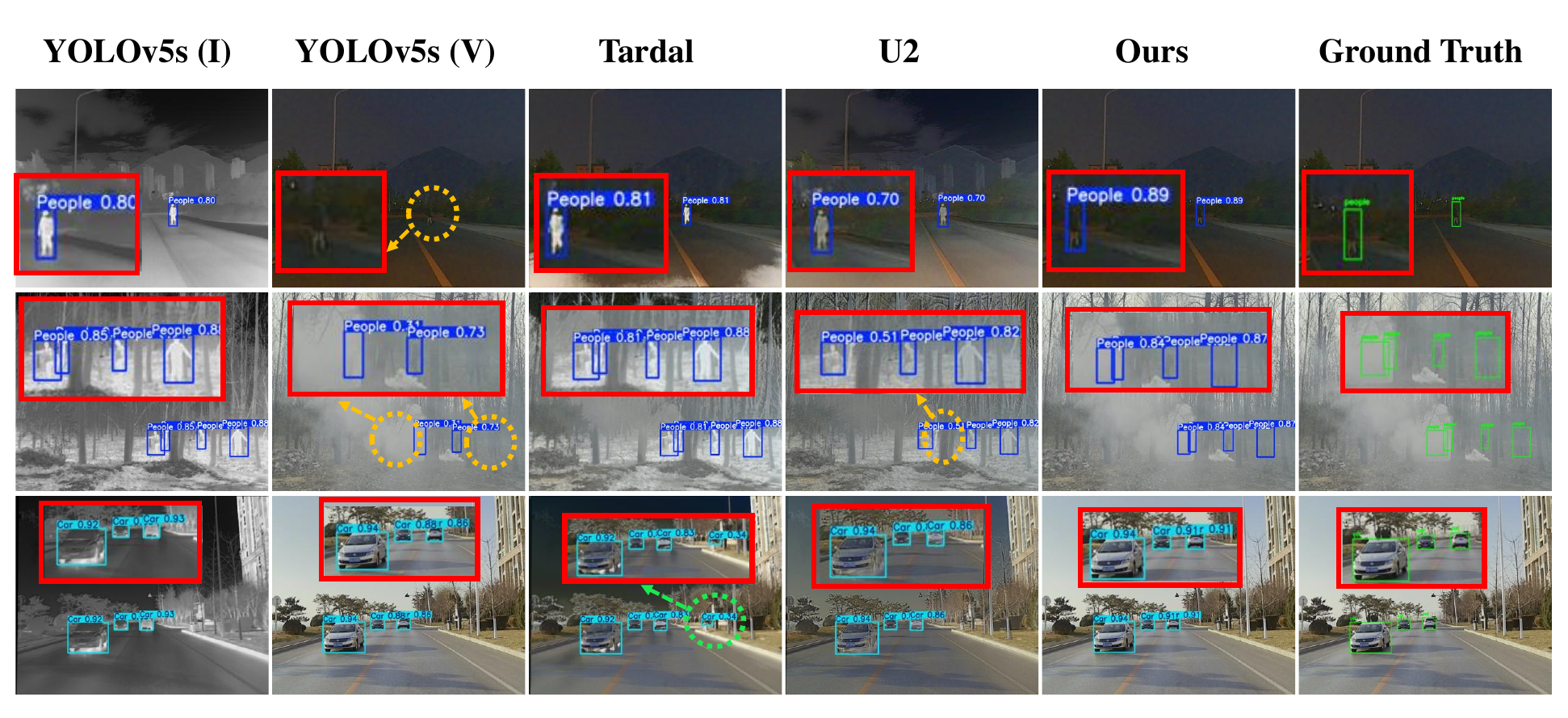} 
    \caption{Comparison of visual object detection results on the M3FD dataset. The figure presents detection outcomes from single-modality, existing multimodal, and the proposed methods. Detection errors are highlighted using dashed circles: \textcolor{green}{green} for false positives, \textcolor{orange}{orange} for missed detections.}
    \label{fig6}
\end{figure*}

\begin{figure}[t]
    \centering
    \includegraphics[width=\columnwidth]{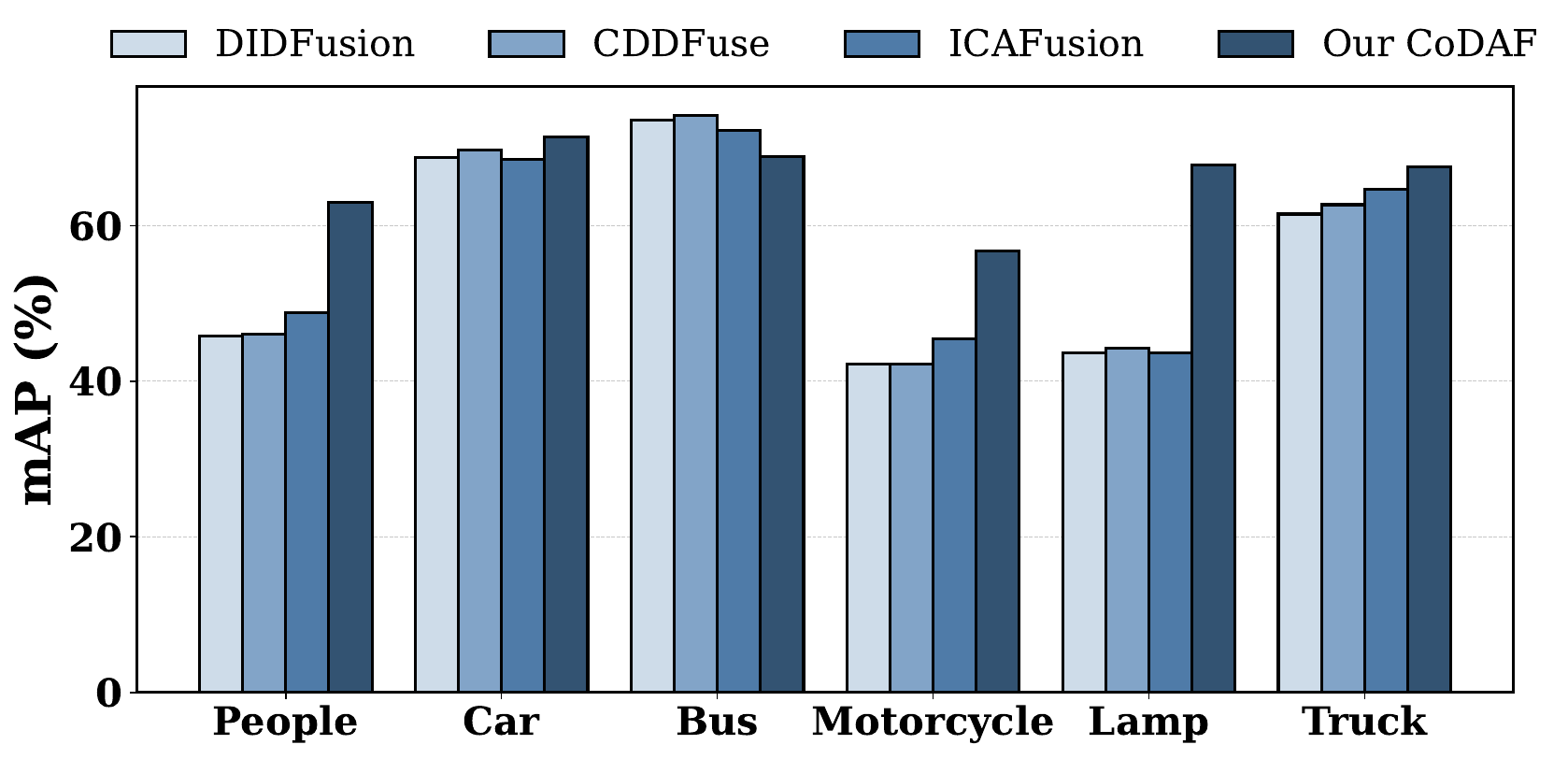} 
    \caption{Comparison of different methods on the $M3FD$ dataset. Our CoDAF achieves higher mAP results through a more effective strategy for cross-modality fusion.}
    \label{fig7}
\end{figure}

Fig.~\ref{fig6} presents visual detection results of the compared methods. CoDAF achieves clearer and more accurate predictions, benefiting from the proposed OSA and DAFM modules, which strengthen cross-modal feature interaction during fusion. These results further demonstrate the robustness and effectiveness of our approach in complex multimodal detection scenarios.

\begin{table*}[ht]
\centering
\caption{Comparisons With State-of-the-Art Methods on the $M3FD$ Dataset among all the image fusion methods + detector (i.e. YOLOv5s). \textbf{Bold}/\underline{Underline} Indicate the Best/Second-Best Results. The terms `Vi' and `Ir' refer to the visible and infrared modalities, respectively.}.
\begin{tabular}{l|c|cc|cccccc|cc}
\toprule
\text{Method} & \text{Public.} & \text{Vi} & \text{Ir} & \text{People} & \text{Car} & \text{Bus} & \text{Motorcycle} & \text{Lamp} & \text{Truck}  & \text{mAP@.5} & \text{mAP@.5:.95} \\
\midrule
YOLOv5s~\cite{yolov5s}   & - & \checkmark & $\times$ & 38.1 & 69.4 & 75.5 & 44.4 & 44.8 & 63.2 & 86.3 & 55.9 \\
\midrule
YOLOv5s~\cite{yolov5s}  & - & $\times$ & \checkmark   & 49.3 & 67.1 & 72.9 & 35.8 & 43.6 & 61.6 & 85.3 & 55.1 \\
\midrule
DIDFusion~\cite{didfuse}   & [arXiv '20] & \checkmark  & \checkmark & 45.8 & 68.8 & 73.6 & 42.2 & 43.7 & 61.5 & 86.2 & 56.2 \\
U2Fusion~\cite{u2fusion}    & [TPAMI '20] & \checkmark  & \checkmark & 47.7 & \textbf{70.1} & 73.2 & 43.2 & 44.6 & 63.9 & 87.1 & 57.1 \\
PIAFusion~\cite{piafusion} & [IF '22] & \checkmark  & \checkmark & 46.5 & 69.6 & 75.1 & 45.4 & 44.8 & 61.7 & 87.3 & 57.2 \\
SwinFusion~\cite{swinfusion} & [JAS '21] & \checkmark  & \checkmark & 44.5 & 68.5 & 73.3 & 42.2 & 44.4 & 63.5 & 85.8 & 56.1 \\
Tardal~\cite{tardal}    & [CVPR '22] & \checkmark  & \checkmark & 49.8 & 65.4 & 69.5 & 46.6 & 43.7 & 61.1 & 86.0 & 56.0 \\
MetaFusion~\cite{metafusion} & [CVPR '23] & \checkmark & \checkmark & 48.4 & 66.7 & 70.5 & 49.1 & 46.4 & 59.9 & 86.7 & 56.8 \\
CDDFuse~\cite{cddfuse}   & [CVPR '23]  & \checkmark  & \checkmark & 46.1 & \underline{69.7} & 74.2 & 42.2 & 44.2 & 62.7 & 87.0 & 56.5 \\
\midrule
CFT~\cite{cft}      & [arXiv '21] & \checkmark  & \checkmark & \textbf{52.0} & 68.2 & \textbf{79.2} & 49.9 & 45.2 & \textbf{69.6} & \underline{89.8} & \underline{60.7} \\
ICAFusion~\cite{icafusion}  & [PR '23] & \checkmark  & \checkmark & 48.8 & 68.5 & 72.3 & 45.5 & 43.6 & 64.7 & 87.4 & 57.2 \\
E2E-MFD~\cite{e2e}     & [NeurIPS '24]  & \checkmark & \checkmark & \underline{51.0} & 67.9 & 69.4 & \underline{50.2} & \underline{48.7} & 61.6 & 87.9 & 58.1 \\
\midrule
\rowcolor[HTML]{b8d6e0} Ours  & - & \checkmark  & \checkmark & 46.8 & 69.4 & \underline{78.0} & \textbf{54.2} & \textbf{50.7} & \underline{68.4} & \textbf{90.8} & \textbf{61.2} \\
\bottomrule
\end{tabular}
\label{tab2}
\end{table*}

\subsection{Ablation Study}

To evaluate the effectiveness of each component within the proposed CoDAF framework, we conduct a series of ablation studies on the DroneVehicle dataset.

\begin{table}[t]
    \centering
    \caption{Ablation study on DroneVehicle dataset. `O' denotes OSA, `M' denotes MAGN, `DA' denotes DACM and `D' denotes DAFM.}
    \setlength{\tabcolsep}{2.5 mm}
    \resizebox{\linewidth}{!}{ 
    \begin{tabular}{lcccc}
        \toprule
        \multirow{2}{*}{Method} & \multirow{2}{*}{OSA} & \multicolumn{2}{c}{DAFM} &  \multirow{2}{*}{mAP@.5}  \\
        \cmidrule(lr){3-4}
        & & MAGN & DACM & \\
        \midrule
        Baseline &  &  &   & 73.9 \\
        Baseline+O & \checkmark &  &  & 77.8 \\
        Baseline+M &  &\checkmark  &  & 77.3 \\
        Baseline+DA &  & &\checkmark  & 75.4 \\
        Baseline+D & &\checkmark  & \checkmark  & 77.7 \\
        \rowcolor[HTML]{b8d6e0} CoDAF & \checkmark & \checkmark & \checkmark & \textbf{78.6} \\
        \bottomrule
    \end{tabular}
    }
\label{tab3}
\end{table}

\subsubsection{\textbf{Effectiveness of the OSA}}

To address semantic inconsistency caused by cross-modal misalignment, we introduce the OSA, which progressively aligns features from infrared and visible modalities in both spatial and semantic domains. As presented in Table~\ref{tab3}, integrating OSA into the baseline yields a notable performance gain of 5.9\% mAP, highlighting the efficacy of the proposed alignment strategy in weakly aligned scenarios. We further perform component-wise ablation to investigate the contribution of each design within OSA. As illustrated in Table~\ref{tab4}, removing contrastive learning supervision (w/o CL) causes a 0.5\% mAP degradation, which confirms the importance of enforcing semantic consistency to guide accurate cross-modal alignment. Furthermore, we evaluated the impact of the spatial attention guidance source by replacing the IR-based spatial attention map ($\mathcal{M}$) with one generated from the visible modality ($\mathcal{M}_v$). As reported in Table~\ref{tab5}, this modification results in a drop of 1.5\% in mAP, suggesting that the IR modality offers more reliable and distinctive spatial cues, especially in low-illumination or complex environments. Thus, it serves as a more robust reference for spatial attention generation.

\begin{table}[ht]
\centering
\caption{Ablation study on contrastive loss.}
\setlength{\tabcolsep}{2.5 mm}
\renewcommand{\arraystretch}{1.1}
\resizebox{\linewidth}{!}{ 
\begin{tabular}{l|ccccc|c}
\toprule
&\text{Car} & \text{Truck} & \text{Fre.} & \text{Bus} & \text{Van} & \text{mAP@.5} \\
\midrule
$CL_{w/o}$  & 93.1 & 77.4 & 65.8 & 90.2 & 64.0 & 78.1 \\
$CL$  & \textbf{93.5} & \textbf{77.9} & \textbf{66.2} &  \textbf{90.5} & \textbf{65.0} & \textbf{78.6} \\
\bottomrule
\end{tabular}
}
\label{tab4}
\end{table}

\subsubsection{\textbf{Anaysis of the Alignment Loss}}

\begin{figure}[htbp]
    \centering
    \includegraphics[width=\columnwidth]{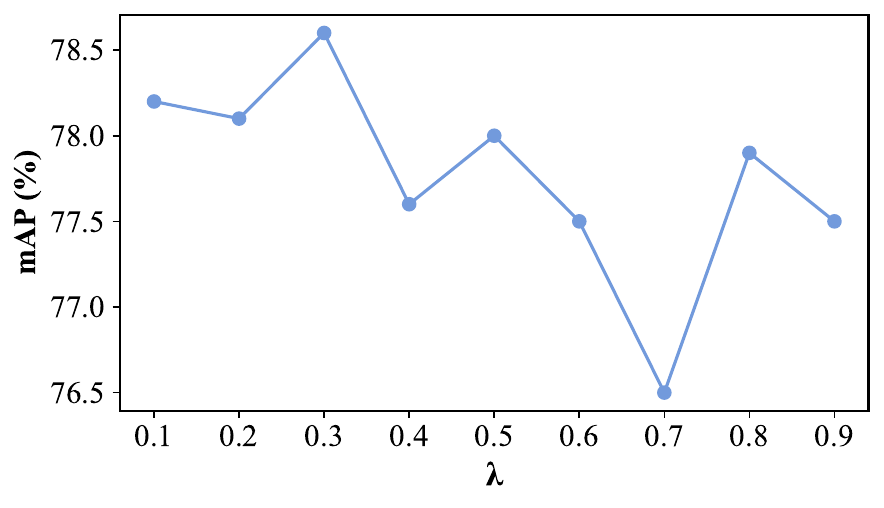} 
    \caption{Ablation study on different value of the $\lambda$.}
    \label{fig8}
\end{figure}

To investigate the influence of the trade-off parameter $\lambda$ on the total alignment loss ($\mathcal{L}_{t}$), we conduct an ablation study by varying $\lambda$  from 0.1 to 0.9 with a step of 0.1. As presented in Fig.~\ref{fig8}, the performance initially increases and reaches the best mAP at $\lambda$  = 0.3. A smaller $\lambda$ underemphasizes the alignment constraint, resulting in suboptimal feature matching. Conversely, an excessively large $\lambda$ dominates the training objective and may suppress discriminative information. These results demonstrate that a proper balance between the alignment and detection objectives is crucial to optimal performance.

\begin{table}[t]
\centering
\caption{Ablation study on spatial attention map.}
\setlength{\tabcolsep}{2.5 mm}
\renewcommand{\arraystretch}{1.1}
\resizebox{\linewidth}{!}{ 
\begin{tabular}{l|ccccc|c}
\toprule
&\text{Car} & \text{Truck} & \text{Fre.} & \text{Bus} & \text{Van} & \text{mAP@.5} \\
\midrule
$\mathcal{M}_{v}$  & 92.9 & 75.8 & 63.6 & 90.3 & 63.7 & 77.1 \\
$\mathcal{M}$  & \textbf{93.5} & \textbf{77.9} & \textbf{66.2} &  \textbf{90.5} & \textbf{65.0} & \textbf{78.6} \\
\bottomrule
\end{tabular}
}
\label{tab5}
\end{table}

\subsubsection{\textbf{Anaysis of the DAFM}}

To enhance multimodal feature integration to reduce the issue of modality conflict, we propose the DAFM, which adaptively modulates the contributions of cross-modal features through two complementary branches: MAGM and DACM. As demonstrated in Table~\ref{tab3}, introducing DAFM into our baseline framework leads to a noticeable 3.8\% mAP improvement, demonstrating that adaptive feature re-weighting yields more effective multimodal fusion than simple addition, especially under challenging and varied conditions. To examine the role of each submodule within DAFM, we conduct detailed ablation experiments, also reported in Table~\ref{tab3}. Removing the MAGM branch causes a 2.3\% mAP drop, confirming that dynamic, pixel-wise modulation plays a critical role in suppressing low-confidence regions and mitigating modality inconsistencies. Furthermore, excluding the DACM branch results in a 0.4\% performance degradation, indicating that DACM effectively enhances the discriminative capacity of fused features by leveraging complementary spatial and semantic cues from both modalities. To further evaluate the design superiority of DACM, we replace it with three widely-used attention mechanisms, SE~\cite{se}, CBAM~\cite{cbam}, and CA~\cite{ca}, while keeping the rest of the full CoDAF framework unchanged. As shown in Table~\ref{tab6}, DACM consistently outperforms these alternatives, validating the superiority of DACM in enhancing cross-modal feature fusion through more effective spatial-channel modulation.

\begin{table}[ht]
\centering
\small
\caption{Comparison with different attention mechanisms.}
\begin{minipage}{\linewidth}
\setlength{\tabcolsep}{3.5mm} 
\renewcommand{\arraystretch}{1.1}
\centering
\begin{tabular}{c|ccc}
\toprule
Method & mAP@.5 & mAP@.75 & mAP@.5:95 \\
\midrule
SE~\cite{se} & 75.9 & 56.0 & 49.7 \\
CBAM~\cite{cbam} & 76.7 & 56.1 & 49.8 \\
CA~\cite{ca} & 77.4 & 57.5 & 51.1 \\
\midrule
\rowcolor[HTML]{b8d6e0} Our DACM & \textbf{78.6} & \textbf{58.8} & \textbf{51.9} \\
\bottomrule
\end{tabular}
\end{minipage}
\label{tab6}
\end{table}

\subsubsection{\textbf{Visualization of Feature Map}}
\begin{figure}[t]
    \centering
    \includegraphics[width=\columnwidth]{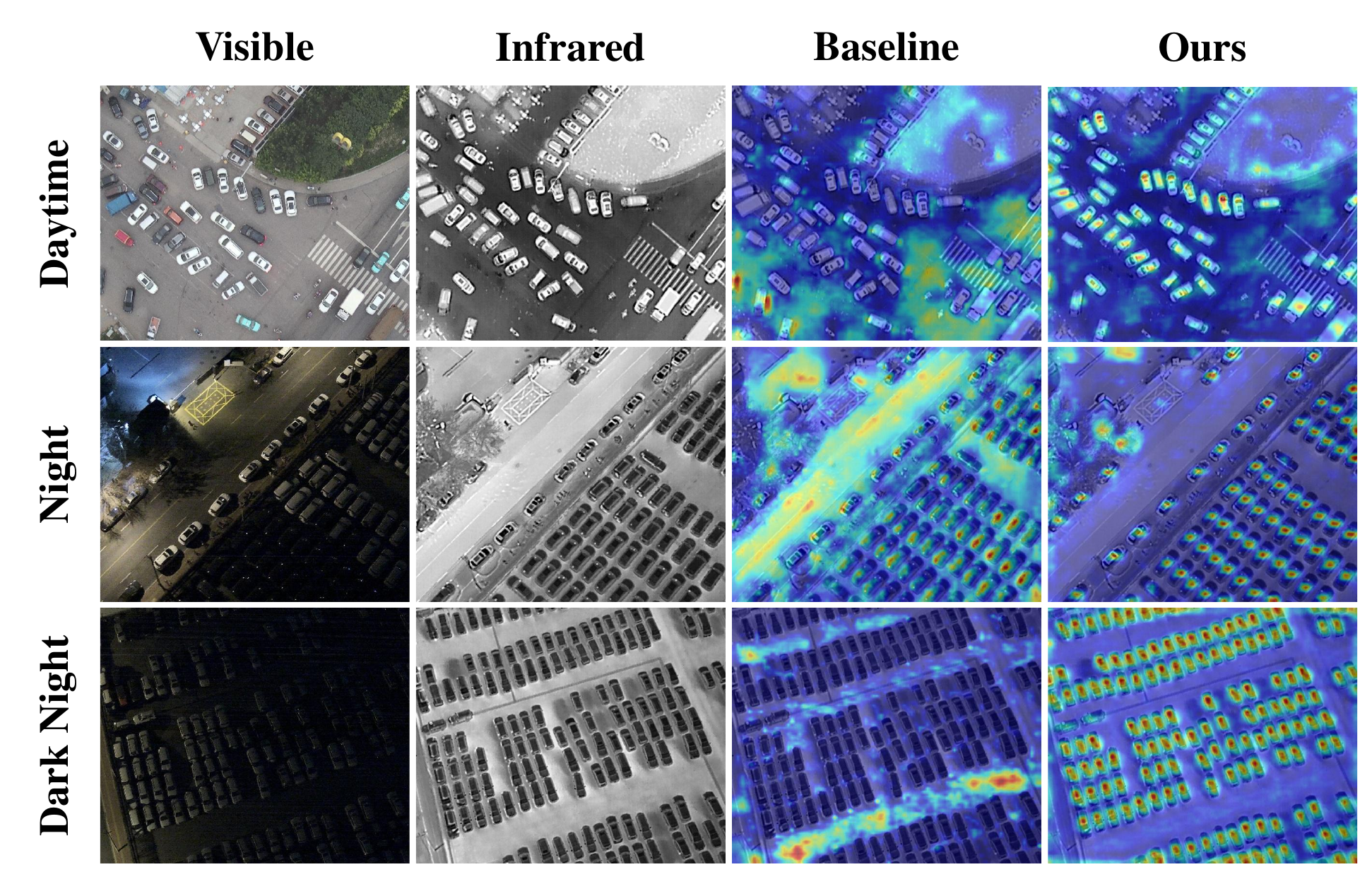} 
    \caption{Feature map visualizations of detection heads from baseline and our CoDAF. From top to bottom, the RGB-IR image pairs correspond to daytime, nighttime, and dark-night scenarios, respectively.}
    \label{fig9}
\end{figure}

As illustrated in Fig.~\ref{fig9}, we present feature map visualizations from the detection heads of the baseline model and our CoDAF-enhanced model across three representative scenarios: daytime, nighttime, and dark night. Each row illustrates a multimodal input pair (visible and infrared) and the corresponding activation maps. Compared to the baseline, which suffers from diffused or background-focused activations, our method consistently generates sharper and more concentrated responses around the target regions.
This visual comparison highlights the effectiveness of our proposed CoDAF module. Notably, even under challenging conditions such as low-light or infrared-dominant scenes, our model maintains strong focus on object areas, thereby improving discriminative capability. Since the baseline shares the same architecture without OSA and DAFM, this comparison clearly demonstrates the overall benefit brought by the CoDAF design in enhancing robust multimodal feature representation.

\subsubsection{\textbf{Effectiveness of OSA and DAFM across stages}}

To investigate the stage-wise effectiveness of our CoDAF design, we performed an ablation study by selectively removing the OSA and DAFM at each stage of the network. As shown in Table~\ref{tab7}, removing these modules in the first stage causes the most significant performance drop (from 78.6\% to 75.4\% mAP@.5), highlighting the importance of early stage alignment and attention for robust multimodal representation. In contrast, removing OSA and DAFM at the second or third stage results in smaller performance degradation, yet still demonstrates the benefit of applying them consistently across the network. Meanwhile, the additional computational cost is minimal, with FLOPs increasing only slightly (e.g., from 217.2G to 224.9G). These results validate the overall efficiency and effectiveness of the proposed OSA and DAFM design at the backbone stages.

\begin{table}[ht]
\centering
\small
\caption{Ablation analysis of OSA and DAFM at different stages.}
\begin{minipage}{\linewidth}
\setlength{\tabcolsep}{1.5mm} 
\renewcommand{\arraystretch}{1.1}
\centering
\begin{tabular}{c|cccc}
\toprule
Method & mAP@.5 & mAP@.75 & mAP@.5:95 & FLOPs \\
\midrule
\rowcolor[HTML]{b8d6e0} our CoDAF & \textbf{78.6} & \textbf{58.8} & \textbf{51.9} & 224.9G \\
\midrule
w/o 1st stage & 75.4 & 50.9 & 46.7 & \textbf{217.2G} \\
w/o 2nd stage & 78.2 & 58.5 & 51.7 & 223.0G \\
w/o 3rd stage & 78.4 & \textbf{58.8} & 51.7 & 224.4G \\
\bottomrule
\end{tabular}
\end{minipage}
\label{tab7}
\end{table}

\begin{table}[ht]
\centering
\small
\caption{Computational cost comparison between CoDAF and SOTA methods.}
\begin{minipage}{\linewidth}
\setlength{\tabcolsep}{1.5mm} 
\renewcommand{\arraystretch}{1.1}
\centering
\begin{tabular}{c|ccccc}
\toprule
Method & Param. & FLOPs & FPS & mAP@.5 & Type \\
\midrule
CIAN~\cite{cian} & - & \textbf{70.36G} & 21.7 & 70.8 & One-stage \\
AR-CNN~\cite{arcnn} & - & 104.3G & 18.2 & 71.6 & Two-stage \\
TSFADet~\cite{tsfadet} & 104.7M & 109.8G & 18.6 & 73.1 & Two-stage \\
CAGTDet~\cite{cagtdet} & - & 120.6G & 17.8 & 74.6 & Two-stage \\
C$^{2}$Former~\cite{c2former} & 100.8M & 89.9G & - & 74.2 & One-stage \\
\midrule
\rowcolor[HTML]{b8d6e0} Our CoDAF & \textbf{67.3M} & 224.9G & \textbf{58.1} & \textbf{78.6} & One-stage \\
\bottomrule
\end{tabular}
\end{minipage}
\label{tab8}
\end{table}

\subsubsection{\textbf{Computational Cost Comparison}}
Table~\ref{tab8} presents a comparison of our CoDAF with some SOTA methods in terms of efficiency and detection performance. CoDAF achieves the best mAP@.5 (78.6\%) while maintaining the lowest parameter count (67.3M), highlighting its lightweight design. Although its FLOPs (224.9G) are relatively high due to the introduced alignment and fusion modules, CoDAF still delivers the highest inference speed (58.1 FPS), surpassing both one-stage and two-stage detectors by a large margin. These results demonstrate that CoDAF offers a favorable trade-off, achieving superior accuracy and real-time performance with minimal parameter overhead. Although CoDAF has higher FLOPs, its RT-DETR-based one-stage architecture ensures fully parallel inference, achieving significantly higher FPS than multi-stage counterparts.

\section{Conclusion}
In this paper, we proposed CoDAF, a unified framework for RGB-IR object detection in UAV-based scenarios, which effectively addresses the challenges posed by weak cross-modal alignment through two dedicated modules. Specifically, the proposed OSA leverages deformable convolutions guided by cross-modal offsets to align features spatially, while enforcing semantic consistency through contrastive learning in a shared semantic space. To further enhance fusion robustness, the DAFM adaptively selects discriminative and spatially reliable features via dual attention mechanisms, mitigating the negative effects of residual misalignment. Extensive evaluations on benchmark datasets verify that CoDAF substantially surpasses existing RGB-IR detection methods under both aligned and weakly aligned conditions.  

Our method also has limitations. On one hand, CoDAF's high computational cost hinders real-time UAV deployment. We will attempt model compression and acceleration techniques to resolve this issue. On the other hand, limited labeled data constrains performance. As future work, we plan to explore semi-supervised or self-supervised learning methods to enhance adaptability in diverse remote sensing environments.

\bibliography{reference}

\newpage

 





\end{document}